\documentclass[runningheads]{llncs}

\usepackage{graphicx}
\usepackage{amsmath,amssymb}
\usepackage{graphics} % for pdf, bitmapped graphics files
\usepackage{epsfig} % so you can include .eps figures
\usepackage{alltt}  % so you can include program text
\usepackage{cite}
\usepackage{listings}
\usepackage{appendix}
\usepackage{algorithm}
\usepackage{algorithmic}
\usepackage{subfig}
\usepackage{float} %subfigs on multiple pages\part{title}
\usepackage{rotating} %,booktabs} %for rotating table
\usepackage{rotfloat}
\usepackage{pdf14}
\usepackage[width=122mm,left=12mm,paperwidth=146mm,height=193mm,top=12mm,paperheight=217mm]{geometry}

\mainmatter
\title{Multi-modal Image Registration for \\ Correlative Microscopy}
\begin{document}
\titlerunning{Multi-modal Image Registration for Correlative Microscopy}

\authorrunning{Tian Cao, Christopher Zach, Marc Niethammer and etc.}

\author{Tian Cao$^1$\and Christopher Zach$^3$\and Shannon Modla$^4$\and Debbie Powell$^4$\and\\
	Kirk Czymmek$^4$\and Marc Niethammer$^{1,2}$}
\institute{$^1$UNC Chapel Hill/$^2$BRIC, $^3$Microsoft Research Cambridge, $^4$University of Delaware}

\maketitle

%\thispagestyle{fancy}

%\MakeTitlePage
%
%\tableofcontents

%\listoffigures % comment out if you have no figures
%\listoftables  % comment out if you have no tables
%
%\clearpage

% Replace each subsection below with the appropriate contents.
% Feel free to add more (sub)sections if you want. Write clearly
% and concisely.

\begin{abstract}

Correlative microscopy is a methodology combining the functionality of light microscopy with the high resolution of electron microscopy and other microscopy technologies. Image registration for correlative microscopy is quite challenging because it is a multi-modal, multi-scale and multi-dimensional registration problem. In this report, I introduce two methods of image registration for correlative microscopy. The first method is based on fiducials (beads). I generate landmarks from the fiducials and compute the similarity transformation matrix based on three pairs of nearest corresponding landmarks. A least-squares matching process is applied afterwards to further refine the registration. The second method is inspired by the image analogies approach. I introduce the sparse representation model into image analogies. I first train representative image patches (dictionaries) for pre-registered datasets from two different modalities, and then I use the sparse coding technique to transfer a given image to a predicted image from one modality to another based on the learned dictionaries. The final image registration is between the predicted image and the original image corresponding to the given image in the different modality. The method transforms a multi-modal registration problem to a mono-modal one. I test my approaches on Transmission Electron Microscopy (TEM) and confocal microscopy images. Experimental results of the methods are also shown in this report.
\\
\end{abstract}

\section{Introduction}
\begin{figure}[hptb]
%\begin{tabular}{c}
\begin{center}
\subfloat[Confocal Microscopic Image]{\includegraphics[scale=0.34]{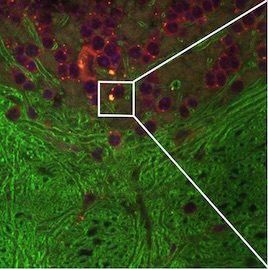}} \;
\subfloat[Resampling of Boxed Region in Confocal Image]{\includegraphics[scale=0.235]{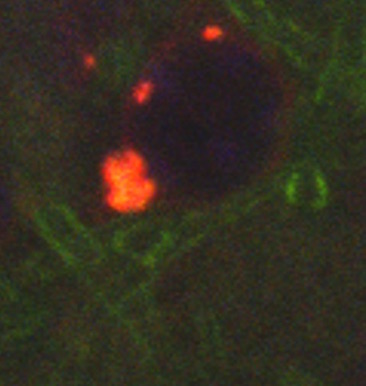}} \;
\subfloat[TEM Image]{\includegraphics[scale=0.4]{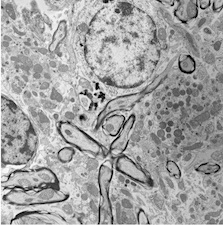}}
%\centering
\end{center}
\caption{Example of Correlative Microscopy. The goal is to align (b) to (c).}
\label{fig:correlative}
\end{figure}

Correlative microscopy is an integration of different microscopy technologies including conventional light, confocal and electron transmission microscopy \cite{caplan2011power}. Correlative microscopic images usually involve linear or non-linear distortions which are caused by the differences between imaging systems and processing steps. Therefore, the first step of most correlative microscopy based applications is to do registration between two or more microscopic images. An example of correlative microscopic images is presented in Fig. \ref{fig:correlative}.

Image registration estimates space transformations between images (to align them) and is an essential part of many image analysis approaches. The registration of correlative microscopic images is very challenging: images should carry distinct information to combine for example knowledge about protein locations (using fluorescence microscopy) with high-resolution structural data (using electron microscopy). However, this precludes the use of simple alignment measures such as the sum of squared intensity differences because intensity patterns do not correspond well or a multi-channel image has to be registered to a gray-valued image. Furthermore, because they operate near or beyond the boundaries of what is measurable, each type of microscopy introduces artifacts into the image that it produces: confocal microscopes convolve the specimen fluorophore distribution with the point-spread-function of their lens system, and scanning electron microscopes produce brighter images near negative surface curvature (as well as volumetric effects), which should ideally be considered when computing alignments.

In this report, I introduce two methods of image registration for correlative microscopy based on fiducials and images respectively. The first method involves automatic landmark based registration. We extract landmarks based on the fiducials and compute the matching landmarks in both images. The transformation matrix is estimated from the corresponding landmarks. We also apply a least-squares matching to the initial alignment of the landmarks to get a better registration result. The second method is inspired by the image analogies approach \cite{hertzmann2001image}. We extend the image analogies using a sparse representation model. %The sparse representation model can generalize the original image analogies method. We also implemented an automatic landmark based image registration method which can be used for further validation of the registration results.

This report is organized as follows. First, I briefly introduce some related work in Sec. 2. Then I describe the automatic landmark based image registration for correlative microscopy in Sec. 3. and the image analogies method with sparse coding and the numerical solutions in Sec. 4. The image registration results are shown in Sec. 5. The conclusion and future work are discussed in Sec. 6.

\section{Related Work}
\subsection{Multi-modal Image Registration for Correlative Microscopy}
One possible solution of image registration for correlative microscopy is to perform landmark-based alignment. Landmarks can be manually specified \cite{kondra2009integration} or automatically extracted\cite{preibisch2010software}. The alignment can be greatly simplified by adding fiducial markers (e.g., some form of beads) which can be identified in the images to be aligned \cite{watanabe2010protein,fronczek2011high}. 

Alternatively one can change the image appearances so that either only image information which is indicative of the desired transformation remains, spurious image information is suppressed, or knowledge about the image formation process is used to convert an image from one modality to another. For example one can register multichannel microscopy images of cells by registering cell segmentations \cite{yang2008nonrigid,yang2011statistical}. The obtained transformations are then applied to the original images.

Since correlative microscopy combines different microscopy modalities, resolution differences between images are common. This poses challenges with respect to finding corresponding regions in the images. If the images are structurally similar (for example when aligning EM images of different resolutions \cite{kaynig2007fully}), standard feature point detectors can be used. 
There are two groups of methods for more general multi-modal image registration \cite{wachinger2010BMVC}. The first set of approaches applies advanced similarity measures, such as mutual information \cite{wells1996multi}. The second group of techniques includes methods that transform a multi-modal to a mono-modal registration. For example, Wachinger introduced entropy images and Laplacian images which are general structural representations \cite{wachinger2010BMVC}. %All these methods require similarity between the distribution of intensities or structures which cannot be directly applied to correlative microscopy.

\subsection{Image Analogies and Sparse Representation}
Image analogies, first introduced in \cite{hertzmann2001image}, have been widely used in texture synthesis.  In this method, a pair of images $A$ and $A'$ are provided as training data, where $A'$ is a ``filtered'' version of $A$. The ``filter'' is learned from $A$ and $A'$ and is later applied to a different image $B$ in order to generate an ``analogous'' filtered image. 

For multi-modal image registration, this method can be used to transfer a given image from one modality to another using the trained ``filter''. Then the multi-modal image registration problem simplifies to a mono-modal one. However, since this method uses a nearest neighbor (NN) search of the image patch centered at each pixel, the resulting images have boundary effects and are usually noisy because the $L_2$ norm based NN search does not preserve the local consistency well (see Fig.\ref{fig:imageanalogy} (d)) \cite{hertzmann2001image}. This problem can be partially solved by a multi-scale search and a coherence search which enforce local consistency among neighboring pixels, but an effective solution is still missing. We introduce a sparse representation model to address this problem.

Sparse representation is a powerful model for representing and compressing high-dimensional signals \cite{wright2010sparse}. It represents the signal with sparse combinations of some fixed bases. Efficient algorithms based on convex optimization or greedy pursuit are available for computing such representations \cite{bruckstein2009sparse}.

%\section{Methods}
%\subsection{Vessel Enhancement Filtering}
\section{Automatic Landmark based Image Registration}

From Sec. 2.1, two common solutions of image registration for correlative microscopy are relying on landmarks and images respectively. Our  methods include an example of each solution. 

If there are fiducials or markers in the images from correlative microscopy and I successfully locate corresponding fiducials, I can estimate the transformation directly from the matching fiducials. However, the locations of the corresponding fiducials are not directly given by the image, so I need to extract the fiducials and find the best matches among them.

We introduce our first algorithm to solve this problem. Our algorithm is based on the algorithm from \cite{serdar2010}, but I use a different and simpler method to locate the fiducials. Whereas in \cite{serdar2010} the author uses a Gaussian distribution to model the shape of fiducials in fluorescence image, I apply a least-squares matching step at the end of the algorithm to refine the initial alignment result. The algorithm description is in Alg. \ref{alg:autolandmark}. \footnotetext[1]{$r=d_1/d_2$ where $d_1$ and $d_2$ are distances of two neighboring landmarks to the landmark $i$ and $d_1\le d_2$.}
%This algorithm is based on the observation of our current test dataset that only similarity transformation involved in the images . 

\begin{algorithm}[htb] %算法的开始
\caption{ Automatic Landmark based Image Registration for Correlative Microscopy.} %算法的标题
\label{alg:autolandmark} %给算法一个标签，这样方便在文中对算法的引用
\begin{algorithmic}[1] %这个1 表示每一行都显示数字
\REQUIRE ~~\ %算法的输入参数：Input
Two images from correlative microscopy: $T$ (Target image) and $S$ (Source image)
\ENSURE ~~\ %算法的输出：Output
%Ensemble of classifiers on the current batch, $E_n$;
Affine transformation $A$ from $S$ to $T$
\STATE Convert $T$ and $S$ to binary images $T_b$ and $S_b$ for fiducials;
\STATE Detect connected components in $T_b$ and $S_b$;
\STATE Calculate the centers of connected components as landmarks in $T_b$ and $S_b$ (to locate the fiducials);
\FOR {each landmark $i$ in $T_b$ and $S_b$}
	\STATE Find its two nearest neighboring landmarks using Euclidean distance;
	\STATE Calculate the ratio $r$ of distances of neighboring landmarks to the landmark $i$\footnotemark[1];
\ENDFOR
\FOR {each landmark $i$ in $S_b$}
	\STATE Find the candidate match $j$ in $T_b$ based on $r$;
	\STATE Calculate the affine transformation matrix $A$ based on the three matching pairs;
	\STATE Apply $A$ to every landmark in $S_b$, calculate the median distance of the nearest points and assign it as the error of the transformation;
	\STATE Record $A$ and its corresponding error;
\ENDFOR
\STATE Find $A_l$ which gives the least median error;%算法的返回值
\STATE Apply a least-squares matching to the corresponding landmarks based on the result of $A_l$ to estimate the refined transformation $A_r$;
\STATE Return $A_r$.
\end{algorithmic}
\end{algorithm}

\begin{figure}[hptb]
%\begin{tabular}{c}
\begin{center}
\subfloat[Source image with landmarks]{\includegraphics[scale=0.3]{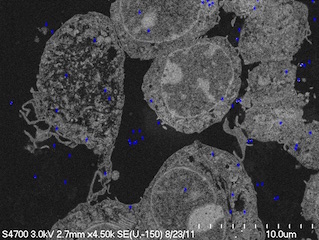}} \;
\subfloat[Target image with landmarks]{\includegraphics[scale=0.4]{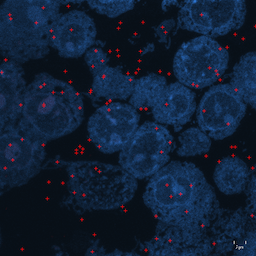}} \\
\subfloat[Composition of transformed source image and target image]{\includegraphics[scale=0.3]{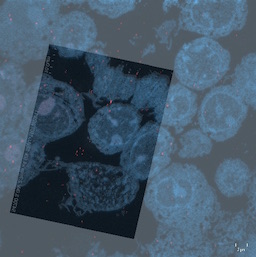}} \;
\subfloat[Registered landmarks]{\includegraphics[scale=0.35]{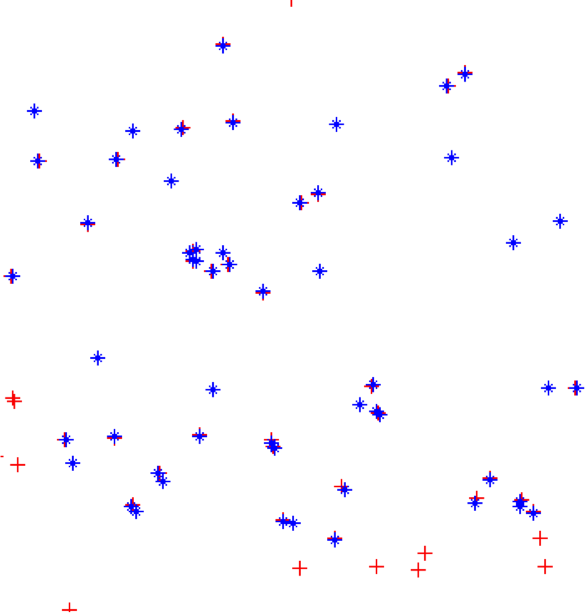}}
%\centering
\end{center}
\caption{Result of Automatic Landmark based Registration}
\label{fig:fiducialresult}
\end{figure}

Theoretically, this algorithm works only for similarity transformations because it finds the corresponding matches of the fiducials based on three nearest landmarks and their distance ratio which is not invariant to shear, nonuniform scale and non-rigid transformations. However, in our test datasets, only weak shear and nonuniform scale transformations occur between correlative microscopic images. Therefore, I can still use this algorithm to estimate the affine transformation. The result of Alg. \ref{alg:autolandmark} is shown in Fig. \ref{fig:fiducialresult}. The quantitative results of Algorithm 1 is shown in Sec. 5.4. The prerequisite of this algorithm is that the fiducials exist in the images which prohibits the application of this algorithm to images without fiducials. Another option is to choose the markers manually, but new errors will be introduced in this process. Since I am also interested in the multi-modal image registration for correlative microscopy without landmarks, I introduce the image analogies method \cite{hertzmann2001image} in Sec. 4 to try to solve this problem.

\section{Image Analogies}

In this section I describe the original image analogies approach of \cite{hertzmann2001image} in Sec. 4.1 and our sparse representation approach in Sec. 4.2.

\subsection{Original Image Analogies Method}

\begin{figure}[hptb]
%\begin{tabular}{c}
\begin{center}
\subfloat[$A$: source TEM image]{\includegraphics[scale=0.21]{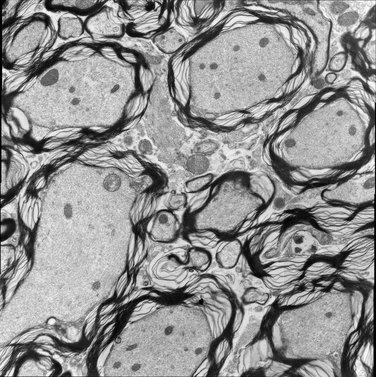}} \;
\subfloat[$A'$: source confocal image]{\includegraphics[scale=0.21]{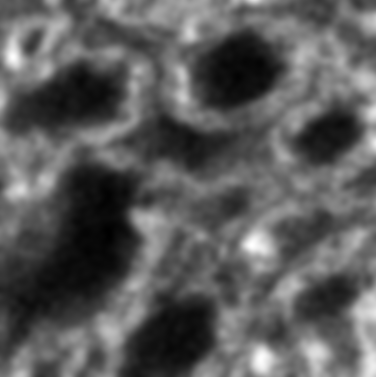}}\;
\subfloat[$B$: input TEM image]{\includegraphics[scale=0.17]{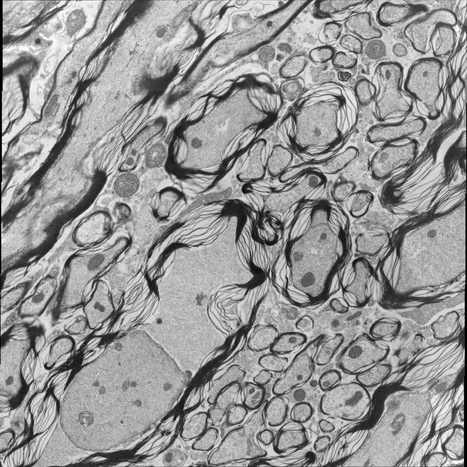}} \;\\
\subfloat[$B'$: output image]{\includegraphics[scale=0.17]{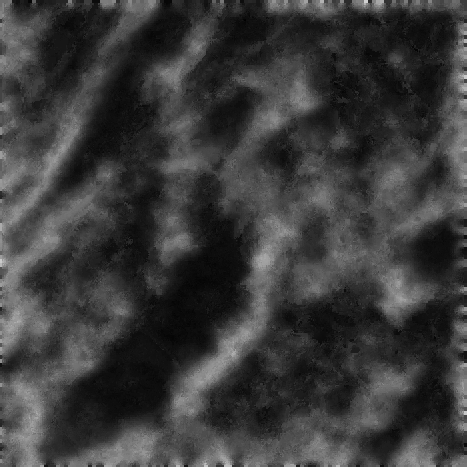}}\;
\subfloat[Real measured confocal image corresponding to B]{\includegraphics[scale=0.17]{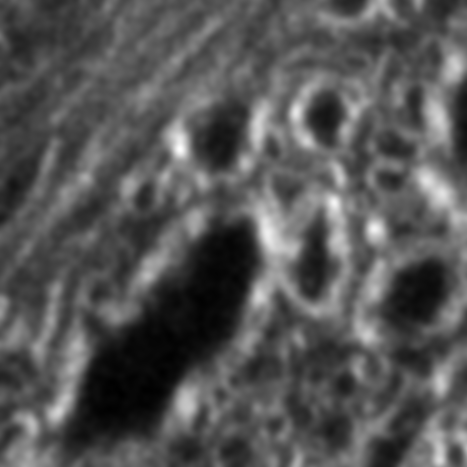}} 
%\centering
\end{center}
\caption{Result of Image Analogy}
\label{fig:imageanalogy}
\end{figure}

The objective for image analogies is to create an image $B'$ from an image $B$ with a similar relation in appearance as a training image set ($A,A'$). 
An example using image analogies \cite{hertzmann2001image} is shown in Fig. \ref{fig:imageanalogy}. %(Fixme: add a figure illustrating how this can help you with the registration) Image analogies are a method to learn the transformation of an image from training examples. A ``filter'' is learned from the training images $A$ and $A'$, and then applied to $B$ to synthesize $B'$. 

The algorithm description in Alg. \ref{alg:imageanalogy}.

\begin{algorithm}[htb]
\caption{ Image Analogies.} 
\label{alg:imageanalogy}
\begin{algorithmic}[1] 
\REQUIRE ~~\ 
%The set of positive samples for current batch, $P_n$;\
%The set of unlabelled samples for current batch, $U_n$;\
%Ensemble of classifiers on former batches, $E_{n-1}$;
\\Training images: $A$ and $A'$;\\
Target image: $B$.
\ENSURE ~~\ %算法的输出：Output
\\Filtered target $B'$.
\STATE Construct Gaussian pyramids for $A$, $A'$ and $B$;
\STATE Generate features for $A$, $A'$ and $B$;
\FOR {each level $l$ starting from coarsest}
	\FOR {each pixel $q\in B_l'$, in scan-line order}
		\STATE Find best matching pixel $p$ of $q$ in $A_l$ and $A_l'$;
		\STATE Assign the value of pixel $p$ in $A'$ to the value of pixel $q$ in $B_l'$;
		\STATE Record the position of $p$. 
	\ENDFOR
\ENDFOR
\STATE Return $B_L'$ where $L$ is the finest level.
\end{algorithmic}
\end{algorithm}

In this algorithm, the features are generated from the raw image patches centered at each pixel and the luminance of color images is used to replace the RGB channels. The most crucial part of this algorithm is to find the best matching pixel $p$ in $A$ and $A'$ of pixel $q$ in $B$ and $B'$. The authors combine an approximated nearest neighbors (ANN) search and a coherence search to find the best match. ANN is an approximated solution of nearest neighbor search in order to balance the accuracy and the time efficiency. However the $L_2$-norm based ANN ignores the local consistency of the image. Thus a coherence search is also applied in finding the best matching pixels. Coherence search is based on the approach by Ashikhmin \cite{ashikhmin2001synthesizing}. The basic idea of coherence search is to favor pixels that are close together to maintain the local consistency. We provide an alternative optimization based approach to image analogies in Sec. 4.2.

\subsection{Sparse Representation Model}

Sparse representation is a technique to reconstruct a signal as a linear combination of a few basis signals from a typically over-complete dictionary. A dictionary is a collection of the basis signals while an over-complete dictionary means the number of signals exceeds the dimension of the signal space. %(fixme: what do you mean over-complete)

Suppose the dictionary $D$ is pre-defined. Sparse representation includes solving an optimization problem \cite{elad2010role} of the form,

\begin{equation}\label{equ:sparsel0}
\begin{aligned}
&\hat{\alpha}=\arg\min_{\alpha}\parallel \alpha \parallel_0,   \\ 
&\text{s.t.}  \parallel x-D\alpha \parallel_2\leq\epsilon,
\end{aligned}
\end{equation}
where $\alpha$ is a sparse vector that explains $x$ as a linear combination of columns in dictionary $D$ with error $\epsilon$ and  $\parallel\cdot\parallel_0$ indicates the number of non-zero elements in the vector $\alpha$. Solving (\ref{equ:sparsel0}) is an NP-hard problem. One possible solution of this problem is based on a relaxation that replaces $\parallel \cdot\parallel_0$ by $\parallel \cdot \parallel_1$, where $\parallel \cdot\parallel_1$ is the $L_1$ norm of a vector, resulting in the following optimization problem \cite{elad2010role}, 

\begin{equation}\label{equ:sparsel1}
\begin{aligned}
&\hat{\alpha}=\arg\min_{\alpha}\parallel \alpha \parallel_1,   \\ 
&\text{s.t.}  \parallel x-D\alpha \parallel_2\leq\epsilon.
\end{aligned}
\end{equation}
The equivalent Lagrangian form of (\ref{equ:sparsel1}) is

\begin{equation}\label{equ:sparsel1lag}
%\begin{aligned}
\hat{\alpha}=\arg\min_{\alpha}\lambda\parallel \alpha \parallel_1  + \parallel x-D\alpha \parallel_2^2,
%\end{aligned}
\end{equation}
where (\ref{equ:sparsel1lag}) is a convex problem and can be solved efficiently \cite{bruckstein2009sparse,boyd2010distributed}.

A more general sparse representation model optimizes both $\alpha$ and the dictionary $D$,

\begin{equation}\label{equ:generalsparse}
%\begin{aligned}
\{\hat{\alpha}, \hat{D}\}=\arg\min_{\alpha,D}\lambda\parallel \alpha \parallel_1  + \parallel x-D\alpha \parallel_2^2.
%\end{aligned}
\end{equation}
The optimization problem (\ref{equ:sparsel1lag}) is a sparse coding problem which finds the sparse codes ($\alpha$) to represent  $x$. Generating the dictionary $D$ from a training dataset is called dictionary learning.

\subsection{Image Analogies with Sparse Representation Model}
For the image registration of correlative microscopy, given two training images $A$ and $A'$ from different modalities, I can transform image $B$ to another modality by synthesizing the image $B'$. This converts a multi-modal registration problem to a mono-modal one. Considering the sparse, dictionary-based image denoising/reconstruction problem of the form
\begin{equation}\label{denoising}
E(u,\{\alpha_i\})=\gamma\int\frac{1}{2}(Lu-f)^2 dx + \frac{1}{N}\bigg(\sum_{i=1}^N\frac{1}{2}\parallel R_iu-D\alpha_i\parallel_V^2 + \lambda\parallel \alpha_i \parallel_1\bigg),
\end{equation}
where $u$ is the sought for image reconstruction, $f$ is the given (potentially noisy) image, $D$ is the dictionary, $\{\alpha_i\}$ are the patch coefficients, $R_i$ selects the i-th patch from the image reconstruction $u$, and $\gamma$, $\lambda>0$  are balancing constants, $L$ is a linear operator (for example describing a convolution operation), and the norm is defined as $\parallel x\parallel_v^2=x^TVx$, where $V>0$ is positive definite. This approach can be extended to image analogies by the reformulation 
\begin{equation}\label{denoising2}
\begin{aligned}
E(u^{(1)},u^{(2)},\{\alpha_i\})&=\gamma\int\frac{1}{2}(L^{(1)}u^{(1)}-f^{(1)})^2+\frac{1}{2}(L^{(2)}u^{(2)}-f^{(2)})^2 dx \\
&+ \frac{1}{N}\bigg(\sum_{i=1}^N\frac{1}{2}\parallel R_i
\left(\begin{array}{c} u^{(1)} \\ u^{(2)} \end{array}\right)-
\left(\begin{array}{c} D^{(1)} \\ D^{(2)} \end{array}\right)\alpha_i\parallel_V^2 
+\lambda\parallel \alpha_i \parallel_1\bigg),
\end{aligned}
\end{equation}
where I have a set of two images $\{f^{(1)},f^{(2)}\}$, their reconstructions $\{u^{(1)},u^{(2)}\}$ and corresponding dictionaries $\{D^{(1)},D^{(2)}\}$. Note that there is only one set of coefficients $\alpha_i$ per patch, which indirectly relates the two reconstructions. See \cite{bruckstein2009sparse,elad2010sparse,wright2010sparse} for an overview on sparse coding and dictionary learning .

Patch-based (non-sparse) denoising has also been proposed for the denoising of fluorescence microscopy images \cite{boulanger2010patch}.
A conceptually similar approach using sparse coding and image patch transfer has been proposed to relate different MR images in \cite{roy2011compressed}. However, this approach does not address dictionary learning, nor is any spatial consistency considered in the sparse coding stage. In our approach dictionaries are learned jointly.

\subsection{Sparse Coding}
Assuming that the two dictionaries $\{D^{(1)},D^{(2)}\}$ are given, the objective is to minimize (\ref{denoising}). However, unlike for the image denoising case of (\ref{denoising2}) for image analogies only one of the images $f^{(1)}$ is given and I am seeking a reconstruction of both, a denoised version of $u^{(1)}$ and $f^{(1)}$ as well as the corresponding analogous denoised image $u^{(2)}$ (without the knowledge of $f^{(2)}$). Hence, for the sparse coding step, (\ref{denoising}) simplifies to
\begin{equation}\label{imganalogy}
\begin{aligned}
E(u^{(1)},u^{(2)},\{\alpha_i\})=&\gamma\int\frac{1}{2}(L^{(1)}u^{(1)}-f^{(1)})^2 dx \\ &+\frac{1}{N}(\sum_{i=1}^N\frac{1}{2}\parallel R_i
\left(\begin{array}{c} u^{(1)} \\ u^{(2)} \end{array}\right)-
\left(\begin{array}{c} D^{(1)} \\ D^{(2)} \end{array}\right)\alpha_i\parallel_V^2  \\
&+\lambda\parallel \alpha_i \parallel_1),
\end{aligned}
\end{equation}
which amounts to a denoising of $f^{(1)}$ inducing a denoised reconstruction of an unknown image $u^{(2)}$. Since the problem is convex (for given dictionaries), I can compute a globally optimal solution. See Sec. 4.6 for a numerical solution approach.
\subsection{Dictionary Learning}
Learning the dictionaries $\{D^{(1)},D^{(2)}\}$ is significantly more challenging than the sparse coding. Here, given sets of training patches $\{p_i^{(1)},p_i^{(2)}\}$ I want to estimate the dictionaries themselves as well as the coefficients $\{\alpha_i\}$ for the sparse coding. Hence, the problem becomes nonconvex (bilinear in $D$ and $\alpha_i$). The standard solution approach \cite{elad2010sparse, kreutz2003dictionary, mairal2008learning} is alternating minimization, i.e., solving for $\alpha_i$ keeping $\{D^{(1)},D^{(2)}\}$ fixed and vice versa. There are two cases: (i) a local invertible operator $L$ (or the identity) and (ii) a non-local operator L (for example blurring due to convolution for a signal with the point spread function of a microscope). In the former case I can assume that the training patches are unrelated, non-overlapping patches and I can compute local patch estimates $\{p^{(1)},p^{(2)}\}$ directly by locally inverting the operator $L$ for the given measurements $\{f^{(1)},f^{(2)}\}$ for each patch. In the latter case, I need to consider full images, because the linear operator may have much further spatial extent than the considered patch size (for example for convolution) and can therefore not easily be inverted patch by patch. The non-local case is significantly more complicated, because the dictionary learning step needs to consider spatial dependencies between patches.

We only consider local dictionary learning here with $L$ and $V$ set to identities. We assume that the training patches $\{p^{(1)},p^{(2)}\}=\{f^{(1)},f^{(2)}\}$ are unrelated, non-overlapping patches. Then the dictionary learning problem decouples from the image reconstruction and requires minimization of
\begin{equation}\label{dictionarylearning}
\begin{aligned}
E_d(D,\{\alpha_i\})& = \sum_{i=1}^N\frac{1}{2}\parallel
\left(\begin{array}{c} f_i^{(1)} \\ f_i^{(2)} \end{array}\right)-
\left(\begin{array}{c} D^{(1)} \\ D^{(2)} \end{array}\right)\alpha_i\parallel^2 + \lambda\parallel \alpha_i \parallel_1 \\
&=\sum_{i=1}^N\frac{1}{2}\parallel f_i-D\alpha_i\parallel^2
+\lambda\parallel \alpha_i\parallel_1.
\end{aligned}
\end{equation}
Thus the image analogy dictionary learning problem is identical to the standard one for image denoising. The only difference is a change in dimension for the dictionary and the patches (which are stacked up for the corresponding image sets). 
 
\subsection{Numerical Solution}
We use a proximal method for the numerical solution of the optimization problems for sparse encoding and dictionary learning \cite{boyd2010distributed,combettes2011proximal}. Proximal methods \cite{combettes2011proximal} have also been used for structured sparse learning \cite{chen2010efficient, mairal2011convex} and hierarchical learning \cite{jenatton2010proximal}. To reduce computational effort for dictionary learning and to make the approaches scalable to millions of patches, online learning methods have been proposed \cite{mairal2009online}.

We use the simultaneous-direction method of multipliers (SDMM) \cite{combettes2011proximal,boyd2010distributed} which allows us to simplify the optimization problem, by breaking it into easier subparts. To apply SDMM, I write the general dictionary learning/image analogy problem as

\begin{equation}\label{eq:SDMM}
\begin{aligned}
E&=\overbrace{\frac{\gamma_1}{2}\parallel v^{(1)}-f^{(1)}\parallel_2^2}^{:=f_D^{(1)}(v^{(1)})}+ \overbrace{\frac{\gamma_2}{2}\parallel v^{(2)}-f^{(2)}\parallel_2^2}^{:=f_D^{(2)}(v^{(2)})} \\ 
&+\frac{1}{N} \bigg( \sum_{i=1}^N \overbrace{\frac{1}{2}\parallel 
\left(\begin{array}{c} v_i^{(1)} \\ v_i^{(2)} \end{array}\right)-\left(\begin{array}{c} w_i^{(1)} \\ w_i^{(2)} \end{array}\right) \parallel_V^2}^{:=f_i^{(p)}\left(\left(\begin{array}{c} v_i^{(1)} \\ v_i^{(2)} \end{array}\right),\left(\begin{array}{c} w_i^{(1)} \\ w_i^{(2)} \end{array}\right)\right) \text{or } \bar{f}_i^{(p)} \left(\begin{array}{c} w_i^{(1)} \\ w_i^{(2)} \end{array}\right)} + \overbrace{\lambda \parallel q_i\parallel_1}^{:=f_i^{(s)}(q_i)} \bigg) \\
&+\underbrace{\frac{\gamma_\alpha}{2}\parallel q\parallel_2^2}_{:f^\alpha(q)},\\
& \text{s.t.} \begin{cases} v^{(1)}=L^{(1)}u^{(1)}\\ v^{(2)}=L^{(2)}u^{(2)}\\ v_i^{(1)}=R_iu^{(1)}\\ v_i^{(2)}=R_iu^{(2)} \end{cases} \text{and }  \begin{cases} w^{(1)}=D^{(1)}\alpha\\ w^{(2)}=D^{(2)}\alpha\\ q_i=W_i\alpha_i\\ q=W\alpha\end{cases},
\end{aligned}
\end{equation}
where I introduced separate copies of the transformed image reconstructions $u^{(1)}$ and $u^{(2)}$ as well as of the patch coefficients and $\alpha$ denotes the stacked up coefficients of all patches (which allows imposing spatial coherence onto the $\alpha_i$ through $W$ if desired). The two alternatives for the patch term, $f_i^{(p)}$ and $\bar{f}_i^{(\hat{p})}$ denote cases where the patches are jointly and not jointly estimated respectively. Therefore, following \cite{combettes2011proximal} I can use an SDMM algorithm which is described in Alg. \ref{alg:basicsdmm}. % where $prox$ is the proximity operator defined in appendix B.

\begin{algorithm}[htb]
\caption{ Basic SDMM algorithm} 
\label{alg:basicsdmm}
\begin{algorithmic} 
\REQUIRE ~~\
$\sigma < 0$
\ENSURE ~~\ 
$x$
\STATE Compute the projection matrix;
\STATE $P=\sum_kL_k^TL_k$;
\WHILE {not converged} 
	\STATE Averaging;
	\STATE $x\leftarrow P^{-1}\sum_kL_k^T(y_k-z_k)$;
	\FOR {$\forall i$} 
		\STATE Create intermediate transformed variable copies;
		\STATE $s_i=L_ix$;
		\STATE Update transformed variable copies;
		\STATE $y_i=prox_{\frac{1}{\sigma}g_i}(s_i+z_i)$ \footnotemark[1];
		\STATE Update dual variables;
		\STATE $z_i\leftarrow z_i+s_i-y_i$.
	\ENDFOR
\ENDWHILE
\end{algorithmic}
\end{algorithm}

For the dictionary-based sparse coding I have three sets of transformed variables, $u^{(1)}$, $u^{(2)}$ and the $\alpha$ copies. The images may even be of different dimensionalities (for example when dealing with a color and a gray-scale image). In our implementation of Algorithm \ref{alg:basicsdmm}, I use $L_1,L_2 = I$ and $W_i = W = I$.\footnotetext[1]{$prox$ is the proximity operator which is defined in appendix B.}

\subsubsection{Dictionary learning}

The primary objective is to predict images given a source image of a given modality (e.g., predicting a fluorescence image from an electron tomography image). We use a dictionary based approach and hence need to be able to learn a suitable dictionary from the data. The local dictionary approach is based on the assumption of independent training patches, whereas the non-local version requires full image-pairs to be able to properly deal with image transformations such as spatial blurring (e.g., caused by a point spread function of a microscope). 

For local dictionary learning, I use an alternating optimization strategy. Assuming that the coefficients ${\alpha_i}$ and the measured patches $\{p_i^{(1)}, p_i^{(2)}\}$ are given, I can compute the current best least-squares solution for the dictionary as \footnote[2]{Refer to appendix A for more details.}

\begin{equation} %\label{Dictlearning}
D=(\sum_{i=1}^Np_i\alpha_i^T)(\sum_{i=1}^N \alpha_i\alpha_i^T)^{-1}.
\end{equation}

The optimization with respect to the ${\alpha_i}$ terms is slightly more complicated, but follows (for each patch independently) the SDMM algorithm. Since the local dictionary learning approach assumes that patches to learn the dictionary from are given, the only terms remaining from Eq. (\ref{eq:SDMM}) are, $\bar{f}_i^{(p)}$ and $f_i^{(s)}$. Hence the problem completely decouples with respect to the coefficients $\alpha_i$ and I obtain

\begin{equation} %\label{Dictlearning}
E=\frac{1}{N}\left(\sum_{i=1}^N \bar{f}_i^{(p)} \left(\begin{array}{c} w_i^{(1)} \\ w_i^{(2)} \end{array}\right)+f_i^{(s)}(q_i) \right), \text{s.t. } w_i^{(1)}=D^{(1)}\alpha_i, \; w_i^{(2)}=D^{(2)}\alpha_i, \; q_i=\alpha_i. 
\end{equation}

\subsubsection{Sparse coding}

Sparse coding follows the same numerical solution approaches for dictionary learning. However, since the dictionaries are known at the sparse coding stage, no alternating optimization is necessary and I can simply solve for $u^{(1)}$ and $u^{(2)}$ using SDMM. The main difference is that for sparse coding for image analogies the measurement of the second image $f^{(2)}$ is unknown. Hence, $f_D^{(2)}(v^{(2)})$ is absent from the optimization and the reconstructed $u^{(2)}$ will be its prediction.

\subsection{Use in Image Registration}
For image registration, I (i) reconstruct the ``missing'' analogous image and (ii) consistently denoise the given image to be registered with. If a local estimation confidence is known, it could be used to weight the image similarity measure locally. Given a scaled image of the local standard deviations of the prediction error, $\omega\in[0,1]$ I modulate the standard sum of squared difference (SSD) similarity measure as %and (iii) compute the local estimation variance
\begin{equation} %\label{denoising}
SSD_w(I_0,I_1)=\sum_i w_i(I_0-I_1)_i^2.
\end{equation}

Other similarity measures (such as cross correlation or mutual information) can be modulated similarly. By denoising the target image using the learned dictionary for the target image from the joint dictionary learning step I obtain two consistently denoised images: the denoised target image and the denoised predicted source image.

\section{Results}
%\subsection{Vessel Enhancement Filtering}
\subsection{Data}

Our test data is from the Delaware Biotechnology Institute Bio-Imaging Center, University of Delaware. We have two sets of data for both image registration methods respectively. For automatic landmark based image registration, I have four pairs of 2D correlative Scanning Electron Microscopy (SEM)/confocal images with $100 nm$ gold fiducials in the datasets. The confocal image is the same in the four dataset and the SEM images are from the same area as the confocal image but different views and magnifications. The pixel size in the confocal image is $40nm$.

For image analogies based image registration, I have corresponding TEM/Confocal Microscopic image pairs of mouse brains with corresponding regions highlighted by a box. Using the correlative microscopy technique on the mouse brain, I want to localize specific brain regions associated with Pelizaeus-Merzbacher Disease (PMD) and do quantitative assessment of hypomyelination and demyelination in mice. PMD is one of a group of genetic disorders characterized by progressive degeneration of the white matter of the brain affecting the myelin sheath, the fatty covering that acts as an insulator on nerve fibers in the central nervous system. 

The confocal microscopy images are multichannel color images in our test dataset. The blue channel is based on the blue stain DAPI (a fluorescent stain) which stains the DNA of the cell nucleus and corresponds to dark regions within the nuclei in the TEM. The green channel is based on the stains of the myelin sheats, visible as dark black layers covering the neurons in the TEM images. The red channel is not explicitly stained for and is caused by the auto-fluorescent effect of lipofuscin. The confocal image with RGB channels and its corresponding TEM image are shown in Fig. \ref{fig:confocalRGB}.

Currently I have six pairs of 2D TEM/confocal images with resolutions 582.24 pixels per $\mu m$ and $7.588$ pixels per $\mu m$ respectively ($1 \,\mu m=1 \,micron = 10^{-6}\,m$). The resolution is different between two images and only a small region in the confocal image corresponds to the TEM image.

\subsection{Automatic Landmark based Image Registration}

\begin{figure}[hptb]
%\begin{tabular}{c}
\begin{center}
\subfloat[Confocal Image]{\includegraphics[scale=0.25]{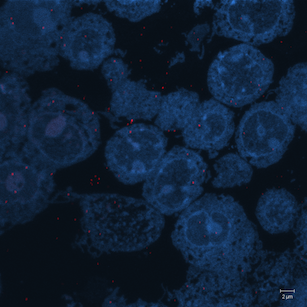}} \
\subfloat[SEM Image in case 1]{\includegraphics[scale=0.25]{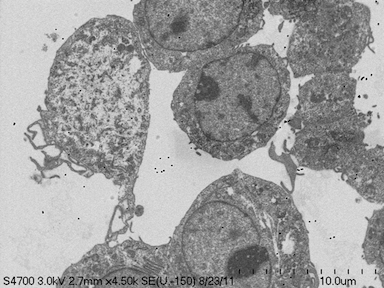}} \
\subfloat[SEM Image in case 2]{\includegraphics[scale=0.8]{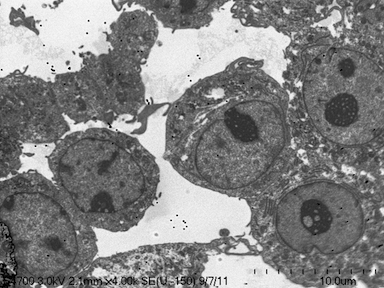}} \
\subfloat[SEM Image in case 3]{\includegraphics[scale=0.8]{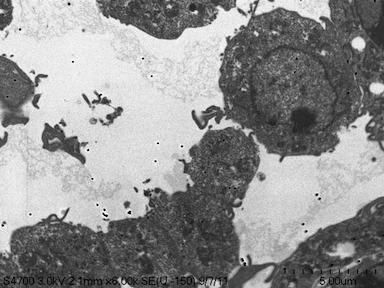}} \
\subfloat[SEM Image in case 4]{\includegraphics[scale=0.8]{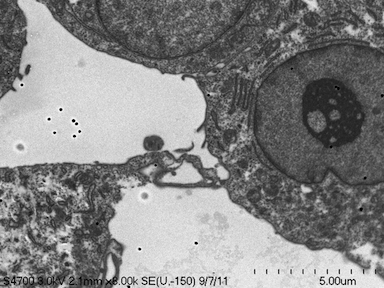}}
%\centering
\end{center}
\caption{Correlative SEM/confocal images}
\label{fig:correlativesem}
\end{figure}

I applied both the proposed automatic landmark based image registration method and the method in \cite{serdar2010}. Fig. \ref{fig:correlativesem} shows the images in the test dataset. I compared the mean absolute errors (MAE) and standard deviations (STD) of the absolute errors on all the corresponding landmarks. The registration results are illustrated in Table \ref{table:landregresult}. The confocal image size is $511 \times 511$ pixels. In ideal case, the MAE and STD should be $0$ (perfect match). Our method improved the registration accuracy in both MAE and STD of the corresponding landmarks.

\begin{table}[ht]
\caption{Landmark based Image Registration Results (in $nm$, pixel size is $40nm$)} % title of Table
\centering  % used for centering table
\begin{tabular}{c c c c c c c} % centered columns (4 columns)
\hline\hline                        %inserts double horizontal lines
 & \multicolumn{2}{c}{Our Method} & \multicolumn{2}{c}{Method in \cite{serdar2010} }\\ [0.5ex] % inserts table
%heading
\hline                  % inserts single horizontal line
case & MAE & STD & MAE & STD \\
\hline
1 & 69.6  & 40.8 & 96.8 & 60 \\ % inserting body of the table
2 & 78.4 & 45.2 & 120.4  & 54.8 \\
3 & 60.8 & 48 & 102.8  & 51.6 \\
4 & 53.2  & 43.2 & 74.8 & 56\\
[1ex]      % [1ex] adds vertical space
\hline %inserts single line
\end{tabular}
\label{table:landregresult} % is used to refer this table in the text
\end{table}

\subsection{Image Analogies based Image Registration}

\subsubsection{Pre-processing}
In the pre-processing step, I extract the corresponding region of the confocal image and resample both confocal and TEM images to an intermediate resolution. The final resolution is 14.52 pixels per $\mu m$, and the image size is about $200 \times 200$ pixels, which is dependent on the original TEM image size. %(Fixme: need also briefly introduce the data used in automatic landmark based registration)

\begin{figure}[hptb]
%\begin{tabular}{c}
\begin{center}
\subfloat[Red channel of confocal image]{\includegraphics[scale=0.2]{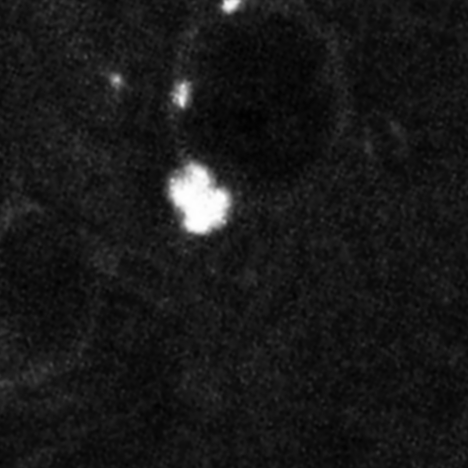}} \
\subfloat[Green channel of confocal image]{\includegraphics[scale=0.2]{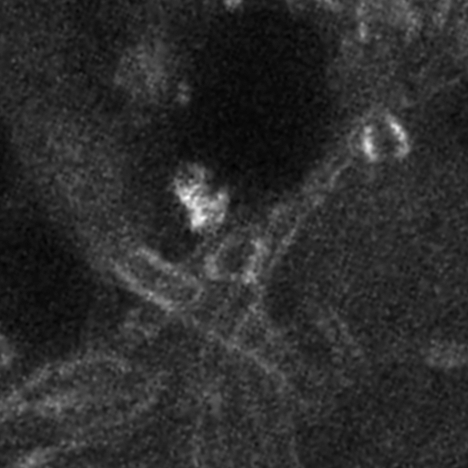}} \
\subfloat[Blue channel of confocal image]{\includegraphics[scale=0.2]{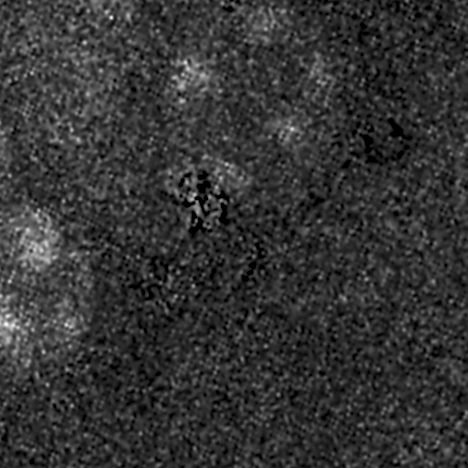}} \
\subfloat[Grayscale of confocal image]{\includegraphics[scale=0.2]{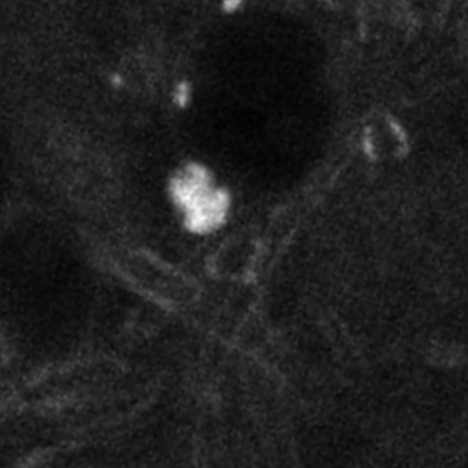}} \
\subfloat[Confocal image]{\includegraphics[scale=0.2]{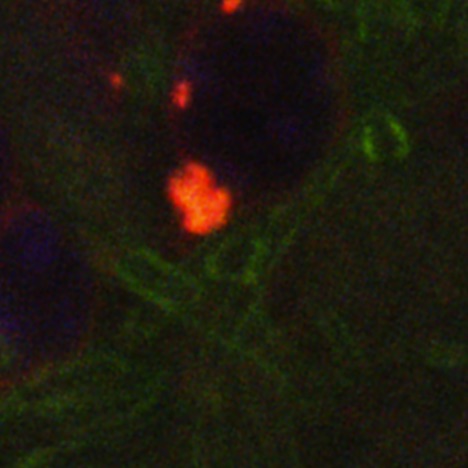}} \
\subfloat[TEM image]{\includegraphics[scale=0.2]{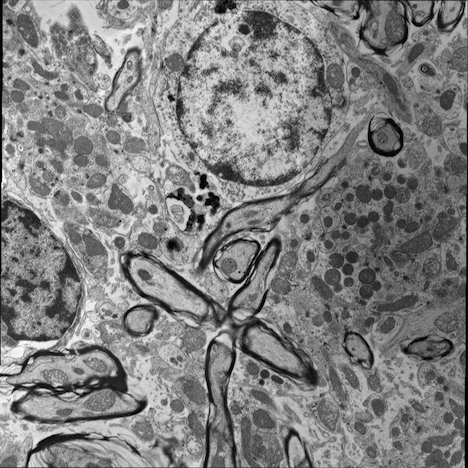}}
%\centering
\end{center}
\caption{Example of Confocal image with RGB channels and TEM image}
\label{fig:confocalRGB}
\end{figure}

From the example in Fig. \ref{fig:confocalRGB}, the blue and red channels are too noisy and contain less information compared to the green channel. We use only the green channel as grayscale image for the registration in our application. The datasets are roughly registered based on manually labeled landmarks with a similarity transformation model.

\subsubsection{Image Analogies Results}
\begin{figure}[hptb]
%\begin{tabular}{c}
\begin{center}
\subfloat[TEM Dictionary]{\includegraphics[scale=0.55]{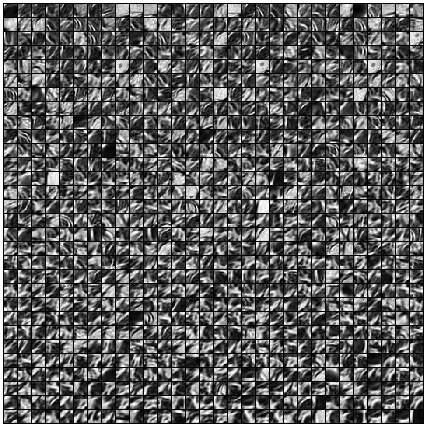}} \;
\subfloat[Confocal Dictionary]{\includegraphics[scale=0.55]{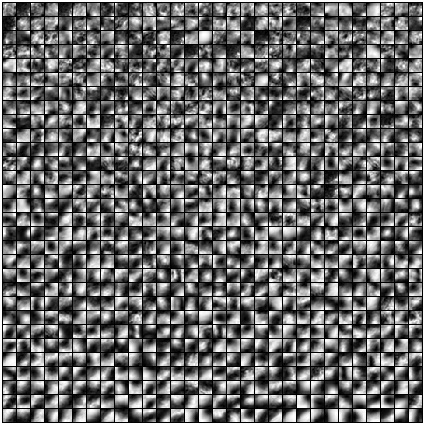}}
%\centering
\end{center}
\caption{Result of Dictionary Learning}
\label{fig:dictlearning}
\end{figure}

We tested the original image analogy method and our proposed method on correlative microscopy images. For each test dataset, I train the dictionaries based on two randomly selected image pairs. We can also train the dictionaries based on more datasets at increased computational cost. In both image analogies methods I use $15 \times 15$ patches, and in our proposed method I randomly sample $10000$ patches and learn $900$ dictionary elements in the dictionary learning phase. We choose $\gamma = 0.01$ and $\lambda = 1$ in (\ref{imganalogy}). The learned dictionaries for both TEM and confocal images are displayed in Fig. \ref{fig:dictlearning}. The image analogies results in Fig. \ref{fig:iaresults} show that our proposed method preserves more local coherence than the original image analogies method.

\begin{figure}[hptb]
%\begin{tabular}{c}
\centering
%\subfloat[TEM Dictionary]{\includegraphics[scale=0.55]{./Fig/temdict.png}}
\subfloat[case 1: TEM image]{\includegraphics[scale=0.37]{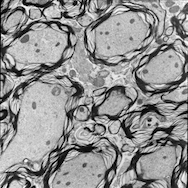}} \
\subfloat[Registered confocal image]{\includegraphics[scale=0.37]{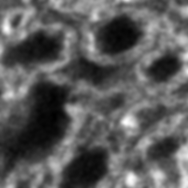}} \
\subfloat[Original IA result]{\includegraphics[scale=0.37]{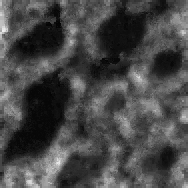}} \
\subfloat[Proposed IA result]{\includegraphics[scale=0.37]{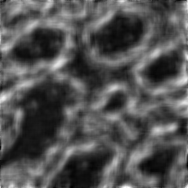}} \\
\subfloat[case 2: TEM image]{\includegraphics[scale=0.3]{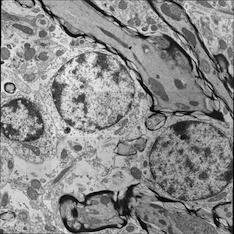}} \
\subfloat[Registered confocal image]{\includegraphics[scale=0.3]{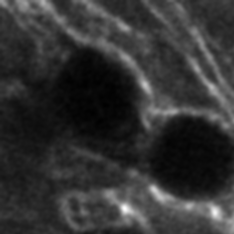}} \
\subfloat[Original IA result]{\includegraphics[scale=0.3]{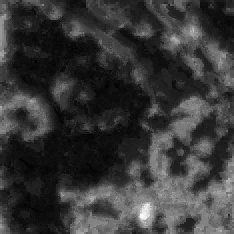}} \
\subfloat[Proposed IA result]{\includegraphics[scale=0.3]{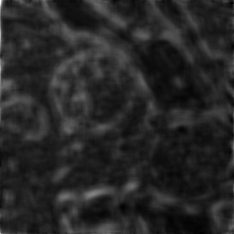}} \\
\subfloat[case 3: TEM image]{\includegraphics[scale=0.3]{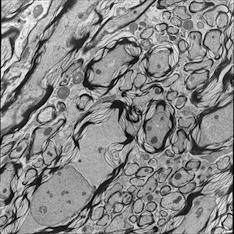}} \
\subfloat[Registered confocal image]{\includegraphics[scale=0.3]{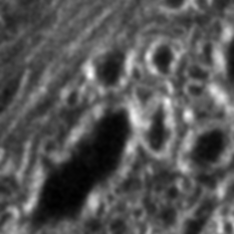}} \
\subfloat[Original IA result]{\includegraphics[scale=0.3]{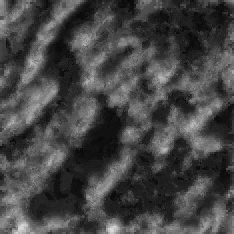}} \
\subfloat[Proposed IA result]{\includegraphics[scale=0.3]{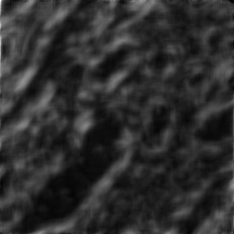}} \\
\subfloat[case 4: TEM image]{\includegraphics[scale=0.3]{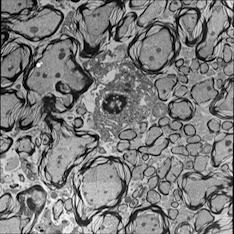}} \
\subfloat[Registered confocal image]{\includegraphics[scale=0.3]{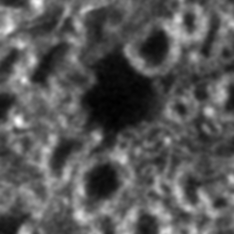}} \
\subfloat[Original IA result]{\includegraphics[scale=0.3]{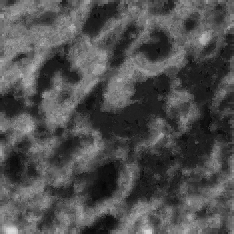}} \
\subfloat[Proposed IA result]{\includegraphics[scale=0.3]{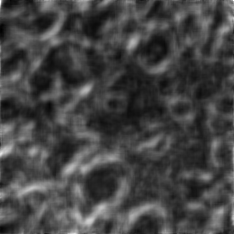}} \\
\caption{Result of Image Analogies (IA)}
\label{fig:iaresults}
\end{figure}

\begin{figure}
\ContinuedFloat
\centering
\subfloat[case 5: TEM image]{\includegraphics[scale=0.3]{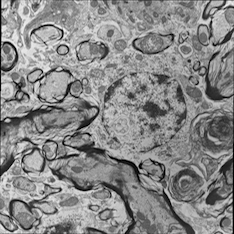}} \
\subfloat[Registered confocal image]{\includegraphics[scale=0.3]{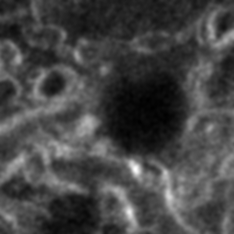}} \
\subfloat[Original IA result]{\includegraphics[scale=0.3]{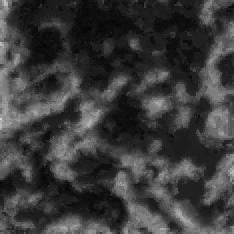}} \
\subfloat[Proposed IA result]{\includegraphics[scale=0.3]{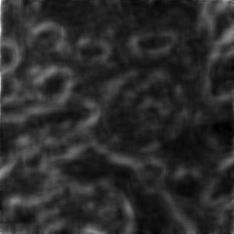}} \\
\subfloat[case 6: TEM image in Case 6]{\includegraphics[scale=0.3]{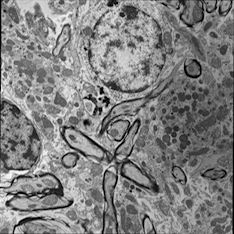}} \
\subfloat[Registered confocal image]{\includegraphics[scale=0.3]{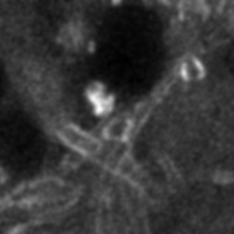}} \
\subfloat[Original IA result]{\includegraphics[scale=0.3]{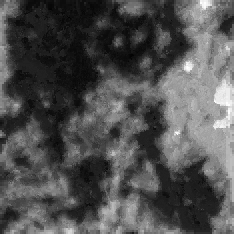}} \
\subfloat[Proposed IA result]{\includegraphics[scale=0.3]{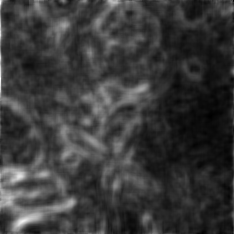}} 
%\centering

\caption{Result of Image Analogies Cont'd}
\label{fig:iaresults}
\end{figure}

\subsubsection{Image Registration Results}

Currently I do not have any gold standard to evaluate the registration results for the correlative microscopy dataset. Thus I manually labeled landmarks and chose about $10\sim 15$ corresponding landmark pairs on each dataset.The image registration results based on both image analogies methods are compared to the landmark based image registration results using the corresponding landmarks. I use both SSD and mutual information (MI) as similarity measure and an affine transformation model. The registration results are shown in Table 2. The landmark based image registration result is the best result I could get based on the transformation model compared to the image analogies based image registration results. I also show the image registration results based on the original TEM and confocal images\footnote[1]{I inverted the grayscale values of original TEM image for SSD based image registration of original TEM/confocal images.}. 

%\newsavebox{\tablebox}
%\begin{lrbox}{\tablebox}\footnotesize
%\begin{table}[ht]
\begin{sidewaystable}
\centering  % used for centering table
\begin{tabular}{c c c c c c c c c c c c c c c} % centered columns (4 columns)
\hline\hline                        %inserts double horizontal lines
& \multicolumn{4}{c}{Our method} & \multicolumn{4}{c}{Original Image analogies} & \multicolumn{4}{c}{Original TEM/Confocal} & \multicolumn{2}{c}{Landmark} \\ [0.5ex] % inserts table
& \multicolumn{2}{c}{SSD} & \multicolumn{2}{c}{MI} & \multicolumn{2}{c}{SSD} & \multicolumn{2}{c}{MI}  &   \multicolumn{2}{c}{SSD} & \multicolumn{2}{c}{MI}  & \\ [0.5ex] 
%heading
\hline                  % inserts single horizontal line
case & MAE  & STD & MAE & STD & MAE  & STD & MAE & STD & MAE  & STD  & MAE  & STD & MAE  & STD\\
\hline
1 & 0.3174 & 0.2698 & 0.3146 & 0.2657 & 0.3119 & 0.2622 & \textbf{0.3036} & 0.2601 & 0.3353 & 0.2519 & 0.5161　& 0.2270 & 0.2705　&　0.1835\\ % inserting body of the table
2 & 0.3912 & 0.1642 & 0.4473 & 0.1869 & \textbf{0.3767} & 0.2160 & 0.4747 & 0.3567 & 2.5420 & 1.6877 & 0.4140 & 0.1780 &　0.3091　&　0.1594\\
3 & 0.4381 & 0.2291 & 0.3864 & 0.2649 & 1.8940 & 1.0447 & 0.4761 & 0.2008 & \textbf{0.4063} & 0.2318 & 0.4078 & 0.2608 &　0.3636　&　0.1746\\
4 & 0.4451 & 0.2194 & 0.4554 & 0.2298 & \textbf{0.4416} & 0.2215 & 0.4250 & 0.2408 & 0.4671 & 0.2484 & 0.4740 & 0.2374 &　0.3823　&　0.2049\\
5 & \textbf{0.3271} & 0.2505 & 0.3843 & 0.2346 & 1.2724 & 0.6734 & 0.4175 & 0.2429 & 0.7204 & 0.3899 & 0.4030 & 0.2519 &　0.2898　&　0.2008\\
6 & 0.7832 & 0.5575 & 0.7259 & 0.4809 & \textbf{0.7169} & 0.4975 & 1.2772 & 0.4285 & 2.2080 & 1.4228 & 0.7183 & 0.4430 &　0.3643　& 0.1435\\ [1ex]      % [1ex] adds vertical space
\hline %inserts single line
\end{tabular}
\label{table:regresult} % is used to refer this table in the text
\caption{Image Registration Results (in $\mu m$, pixel size is 0.069 $\mu m$)} % title of Table
%\end{table}
%\end{lrbox}
%\rotatebox[origin=c]{90}{\usebox{\tablebox}}
\end{sidewaystable}

Table 2 shows that the MI based image registration results are similar among the three methods and also close to the landmark based registration results (best registration results). For SSD based image registration, our proposed method is more robust than the other two methods for the current datasets, for example, using the original image analogies method results in large MAE values in case 3 and case 4 while using the original TEM/confocal images for registration results in large MAE values in case 2 and case 6. While our method does not currently give the best results for all the cases available to us, it appears to be the consistent with results close to the best among all the methods investigated for all cases.

\section{Conclusion}

I have developed two image registration methods for correlative microscopy. The first method is an automatic landmark based method and the second method is based on image analogies with a sparse representation model. The image registration results in Table \ref{table:landregresult} show that our proposed landmark based image registration algorithm can improve accuracy of the registration result using the method in \cite{serdar2010}. However, this method works only on similarity transformations while it is not suitable for shear, nonuniform scale and non-rigid transformations because of the assumption that the distance ratio between three nearest landmarks are invariant to the transformation only holds for similarity transformations but not for general non-rigid transformations. Thus I need to develop a more robust method to estimate the matching landmarks. 
 
The image analogies based method estimates the transformation from one modality to another based on training datasets of two different modalities. The results in Fig. \ref{fig:iaresults} and Table 2 show that both image analogies methods can achieve similar results on image registration for the correlative microscopy datasets while our proposed method can generate ``visually'' better results, i.e. maintaining more local coherence. The image analogies based method is a general method that can be used in other multi-modal image registration problems. 

Our future work includes additional tests and validation on the  datasets from different modalities and the computation of the local estimation variance of the image analogy result. Also, I would like to extend the current image analogies method to 3D images, while the current method focuses only on 2D images.

\appendix
\section*{Appendix}
\subsection*{1 Updating the dictionary}
Assume we are given current patch estimates and dictionary coefficients. The patch estimates can be obtained from an underlying solution step for the non-local dictionary approach or given directly for local dictionary learning. The dictionary-dependent energy can be rewritten as

\begin{displaymath} %\label{Dictlearning}
\begin{aligned}
E_d(D,\{\alpha_i\})&=\sum_{i=1}^N\frac{1}{2}(p_i-D\alpha_i)^T(p_i-D\alpha_i)+\lambda \parallel \alpha_i\parallel_1 \\
&=\sum_{i=1}^N\frac{1}{2}(p_i^Tp_i-p_i^TD\alpha_i-\alpha_i^TD^Tp_i+\alpha_i^TD^TD\alpha_i)+\lambda\parallel\alpha_i\parallel_1.
\end{aligned}
\end{displaymath}
Using the derivation rules \cite{petersen2008matrix}
\begin{displaymath} %\label{Dictlearning}
\frac{\partial a^TXb}{\partial X}=ab^T,   \; \;\; \frac{\partial a^TX^Tb}{\partial X}=ba^T,   \;\;\;   \frac{b^TX^TXc}{\partial X} = Xbc^T+Xcb^T,
\end{displaymath}
we obtain 
\begin{displaymath}
\frac{\partial E_d(D,\{\alpha_i\})}{\partial D} = \sum_{i=1}^N(D\alpha_i-p_i)\alpha_i^T=0.
\end{displaymath}
After some rearranging, we obtain
\begin{displaymath}
D\underbrace{\sum_{i=1}^N\alpha_i\alpha_i^T}_{=A}=\underbrace{\sum_{i=1}^Np_i\alpha_i^T}_{=B}.
\end{displaymath}
If $A$ is invertible and we obtain
\begin{displaymath}
D=(\sum_{i=1}^Np_i\alpha_i^T)(\sum_{i=1}^N\alpha_i\alpha_i^T)^{-1}=BA^{-1}.
\end{displaymath}
If $A$ is not invertible, then the solution is given through the generalized right pseudo-inverse, $A_r^\dag$,
\begin{displaymath}
D=BA^T(AA^T)^{-1}=BA_r^\dagger.
\end{displaymath}

\subsection*{2 Proximity operators}
The proximity operator is defined as 
\begin{displaymath}
\text{prox}_f^L(y)=\arg\min_u f(u)+\frac{1}{2}\parallel Lu-y\parallel_2^2.
\end{displaymath}
For the computations we make extensive use of matrix and vector derivative rules as given in the matrix cookbook \cite{petersen2008matrix}.
\subsubsection*{2.1 $f(x)=c\parallel x\parallel_1$.}
The proximity operator for 
\begin{displaymath}
f(x) = c\parallel x\parallel_1,
\end{displaymath}
is a simple component-wise soft shrinkage operation, i.e.,
\begin{displaymath}
(\text{prox}_{\frac{1}{\sigma}f}(y))_k=\begin{cases} y_k-\frac{c}{\sigma},\; y_k>\frac{c}{\sigma},\\ y_k+\frac{c}{\sigma},\; y_k<-\frac{c}{\sigma},\\ 0,\;\;\;\; \text{otherwise}. \end{cases}
\end{displaymath}
\begin{proof}
We need to find the minimizing argument for
\begin{displaymath}
E(x)=c\parallel x\parallel_1+\frac{\sigma}{2}\parallel x-y\parallel_2^2.
\end{displaymath}
This equation decouples for all $x_i$. Hence for all $x_i$ we need to minimize
\begin{displaymath}
E(x_i)=c\parallel x_i\parallel+\frac{\sigma}{2}(x_i-y_i)^2.
\end{displaymath}
Fox $x_i\geq0$ we obtain
\begin{displaymath}
E(x_i)=cx_i+\frac{\sigma}{2}(x_i-y_i)^2,
\end{displaymath}
and
\begin{displaymath}
\frac{\partial E(x_i)}{\partial x_i}=c+\sigma(x_i-y_i).
\end{displaymath}
Hence $E(x_i)$ is minimal for $x_i=y_i-\frac{c}{\sigma}$. The other cases follow correspondingly.
\end{proof}

\subsubsection*{2.2 $f(x)=\frac{c}{2}\parallel p-Dx \parallel_v^{2}$.}
\begin{proof}
We need to minimize
\begin{displaymath}
\begin{aligned}
E(x)&=\frac{c}{2}\parallel p-Dx\parallel_v^{2}+\frac{\sigma}{2}\parallel x-y\parallel_2^2 \\
&=\frac{c}{2}(p^TVp-p^TVDx-x^TD^TVp-x^TD^TVp+x^TD^TVDx)+\frac{\sigma}{2}\parallel x-y\parallel_2^2.
\end{aligned}
\end{displaymath}
Differentiating the energy yields
\begin{displaymath}
\frac{\partial E}{\partial x}=c(-D^TVp+D^TVDx)+\sigma(x-y)=0.
\end{displaymath}
Hence, $E$ is minimal for 
\begin{displaymath}
x=(\sigma I+cD^TVD)^{-1}(\sigma y+cD^TVp).
\end{displaymath}

\end{proof}

\subsubsection*{2.3 $f(x_1,x_2)=\frac{c}{2}\parallel x_1-Dx_2\parallel_v^{2}$.}
\begin{proof}
We need to minimize
\begin{displaymath}
E(x_1,x_2)=\frac{c}{2}\parallel x_1-Dx_2\parallel_v^{2}+\frac{\sigma}{2}(\parallel x_1 -y_1\parallel_2^2+\parallel x_2-y_2\parallel_2^2).
\end{displaymath}
The derivative with respect to $x_2$ is as in Sec. B.2
\begin{displaymath}
\frac{\partial E}{\partial x_2}=c(-D^TVx_1+D^TVDx_2)+\sigma(x_2-y_2)=0.
\end{displaymath}
The derivative with respect to $x_1$ is 
\begin{displaymath}
\frac{\partial E}{\partial x_1} = c(Vx_1-VDx_2)+\sigma(x_1-y_1)=0.
\end{displaymath}
Writing in matrix form yields
\begin{displaymath}
c\left(\begin{array}{cc} V+\sigma I & -VD \\ -D^TV & D^TVD+\sigma I \end{array}\right)\left(\begin{array}{c}x_1 \\ x_2\end{array}\right)=\sigma \left(\begin{array}{c}y_1 \\ y_2\end{array}\right).
\end{displaymath}
Hence we obtain
\begin{displaymath}
\text{prox}_{\frac{1}{\sigma}f}=\frac{\sigma}{c}\left(\begin{array}{cc} V+\sigma I & -VD \\ -D^TV & D^TVD+\sigma I \end{array}\right)^{-1} \left(\begin{array}{c}y_1 \\ y_2\end{array}\right).
\end{displaymath}
\end{proof}

\bibliographystyle{splncs03}
\bibliography{ref}

\end{document}